%% file: main.tex
\pdfoutput=1

\documentclass[11pt]{article}

\usepackage[final]{acl}

\usepackage{times}
\usepackage{latexsym}
\usepackage{algorithm}

\usepackage[T1]{fontenc}

\usepackage[utf8]{inputenc}

\usepackage{microtype}

\usepackage{inconsolata}
\usepackage{hyperref}
\usepackage{xspace}
\usepackage{booktabs}
\usepackage{multirow}
\usepackage{algpseudocode}
\usepackage{amssymb}

\usepackage{graphicx}

\newcommand{\newscience}{Scipy-M }
\newcommand{\jumptensor}{Tensorflow-M }

\newcommand{\ourmethod}{\textsc{EvoR}\xspace}
\newcommand{\ourdata}{\textsc{EvoR-bench}\xspace}

\usepackage{xcolor,pifont}
\newcommand*\colourcheck[1]{%
  \expandafter\newcommand\csname #1check\endcsname{\textcolor{#1}{\ding{52}}}%
}
\colourcheck{green}
\newcommand*\colourcross[1]{%
  \expandafter\newcommand\csname #1cross\endcsname{\textcolor{#1}{\ding{56}}}%
}
\colourcross{red}
\usepackage{amsmath}
\usepackage{multicol}
\usepackage{fontawesome5} 

\title{\ourmethod: Evolving Retrieval for Code Generation}

\author{
 \textbf{Hongjin Su\textsuperscript{1}},
 \textbf{Shuyang Jiang\textsuperscript{2}},
 \textbf{Yuhang Lai\textsuperscript{2}},
\\
 \textbf{Haoyuan Wu\textsuperscript{1}},
 \textbf{Boao Shi\textsuperscript{1}},
 \textbf{Che Liu\textsuperscript{1}},
 \textbf{Qian Liu\textsuperscript{3}},
 \textbf{Tao Yu \textsuperscript{1}},
\\
\\
 \textsuperscript{1}The University of Hong Kong,
 \textsuperscript{2}Fudan University,
 \textsuperscript{3}Sea AI Lab,
\\
 \small{
   \textbf{Correspondence:} \href{mailto:email@domain}{hjsu@cs.hku.hk}
 }
}

\begin{document}
\maketitle

\input{texts/abstract1}
\input{texts/intro1}
\input{texts/setting}

\input{texts/experiment}

\input{texts/analysis}

\input{texts/related_works}
\input{texts/conclusion}
\input{texts/limitation}


\bibliography{custom}

\input{texts/appendix}

\end{document}

%% file: texts/abstract1.tex
\begin{abstract}
Recently the retrieval-augmented generation (RAG) has been successfully applied in code generation.
However, existing pipelines for retrieval-augmented code generation (RACG) employ static knowledge bases with a single source, limiting the adaptation capabilities of Large Language Models (LLMs) to domains they have insufficient knowledge of.  
In this work, we develop a novel pipeline, \ourmethod, that employs the synchronous evolution of both queries and diverse knowledge bases.
On two realistic settings where the external knowledge is required to solve code generation tasks, we compile four new datasets associated with frequently updated libraries and long-tail programming languages, named \ourdata.
Extensive experiments demonstrate that \ourmethod achieves two to four times of execution accuracy compared to other methods such as Reflexion~\cite{shinn2024reflexion}, DocPrompting~\cite{zhou2023docprompting}, etc. 
We demonstrate that \ourmethod is flexible and can be easily combined with them to achieve further improvement.
Further analysis reveals that \ourmethod benefits from the synchronous evolution of queries and documents and the diverse information sources in the knowledge base.
We hope that our studies will inspire more insights into the design of advanced RACG pipelines in future research.
Our model, code, and data are available at https://arks-codegen.github.io.

\end{abstract}

%% file: texts/intro1.tex
\begin{figure}
    \centering
    \includegraphics[width=\columnwidth]{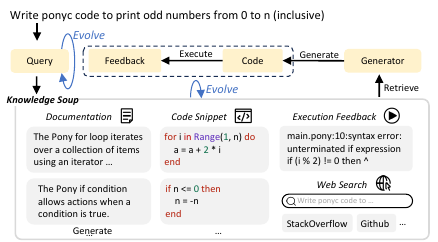}
    \caption{Instead of using a given query to retrieve from a static knowledge base, we design a novel pipeline 
    to dynamically evolve both queries and knowledge soup in retrieval-augmented code generation.
    }
    \label{fig:main}
\end{figure}

\section{Introduction}

The retrieval-augmented generation (RAG) paradigm has raised significant attention due to its efficiency in adapting large language models (LLMs) without training~\citep{guu2020realm,karpukhin-etal-2020-dense,izacard2023atlas,borgeaud2022retro,retrieval-lm-tutorial}. 
Recent research has demonstrated its successful applications in code generation.
They implement the retrieval-augmented code generation (RACG) pipelines either using a given query~\citep{parvez2021retrieval}, or a rewritten version~\citep{jiang2023selfevolve} to retrieve from a static knowledge base with a single type of information, e.g., syntax documentation~\citep{zan2022language,zhou2023docprompting}, code repositories~\citep{zhang-etal-2023-repocoder,shrivastava2023repofusion}, etc.

However, more knowledge sources are potentially helpful to generalization, e.g., web content~\citep{parvez-etal-2021-retrieval-augmented,wang2022execution}, code snippets generated by LLM~\citep{zhang2023repocoder}, etc. 
This information is easily obtained and can enrich knowledge bases, which are shared among all instances of the same task.
Furthermore, the unique characteristic of execution in code generation enables more information collected on-the-fly.
For instance, if a code snippet generated by LLMs is successfully executed without reporting error messages, it is guaranteed to be syntactically correct and can serve as a concrete example to demonstrate the corresponding grammar or function usage.

In this work, we introduce \ourmethod, a novel pipeline that applies synchronous evolution of both queries and documents in RACG.
In the traces of multi-round interactions among retrievers, LLMs and executors, both queries and knowledge bases are updated based on the execution feedback and LLM outputs in every iteration.
This strategic refinement aims to facilitate the extraction of the most pertinent information.
Apart from the given library documentation, we construct a diverse knowledge soup to further integrate the web search content, execution feedback, and code snippets generated by LLMs in the inference time.

To prevent the issue of data leakage associated with large language models pretrained on massive public datasets, and assess \ourmethod under a reliable generalization setting, we compile a new benchmark, \ourdata, comprising four datasets designed to simulate realistic scenarios in RACG. 
Specifically, two of these datasets focus on modifications made to widely-used Python libraries, Scipy and Tensorflow. 
The remaining two datasets simulate the introduction of new grammars, with the help of two less-common programming languages Ring and Pony.
To conduct thorough experiments, we employ both proprietary models, such as ChatGPT~\citep{elmohamed}, and open-source models like CodeLlama~\citep{roziere2023code}.
Experimental results across these four datasets demonstrate that our method yields a significant improvement in the average performance over existing code generation methods.
For example, \ourmethod outperforms DocPrompting~\cite{zhou2023docprompting} by 18.6\% on average using CodeLlama (\S\ref{sec:experiment}). 
Further analysis unveils that both synchronous evolution and diverse sources in knowledge bases are critical to the success of \ourmethod (\S\ref{sec:syn}, \S\ref{sec:knowledge}).
We demonstrate that \ourmethod is flexible to integrate with many other code generation approaches including the agent-based one, e.g., swe-agent, offering further performance enhancement in both \ourdata and existing benchmarks (\S\ref{sec:repo}).
Finally, we showcase \ourmethod is a more effective approach to using tokens, and demonstrates superior results in all levels of token consumption ranging from 4k to 24k (\S\ref{sec:token}).
In summary, our contributions are:
\begin{itemize}
    \item We propose a novel pipeline, \ourmethod, highlighting the complementary strength of synchronous evolution of queries and diverse knowledge bases in RACG.
    \item We compile a new benchmark, \ourdata, on two realistic RACG settings related to frequently updated libraries and long-tail programming languages.
    \item We conduct extensive analyses and find that \ourmethod can be easily combined with existing code generation approaches including agent-based ones to provide further improvements.
\end{itemize}

%% file: texts/setting.tex
\section{Evolving Retrieval}
\label{sec:active_retrieval_in_knowledge_soup}
Given a question $n$ in natural language, the objective of retrieval-augmented code generation is to first retrieve relevant information $K^+$ from external knowledge $K$ and then augment large language models to generate a program $p$ in the target library/programming language, which LLM $M$ is not familiar with.
Distinct from the classical retrieval-augmented generation, which usually focuses on static knowledge bases, we propose synchronous evolution of both queries and diverse knowledge bases.
Intuitively, this helps the retrieval model identify more relevant information and thus improves the quality of LLM generation \cite{shao2023enhancing}.
In this section, we present the process of query evolution (\S\ref{sec:query_formulation}), the knowledge base construction and evolution (\S\ref{sec:knowledge_soup_construction}), the \ourmethod pipeline (\S\ref{sec:active_retrieval_pipelines}) and the collection of \ourdata (\S\ref{sec:datasets}).

\subsection{Query evolution}
\label{sec:query_formulation}
Starting from the given question $n$, we first go through a warmup iteration $i_0$ where $q_0=n$ is used as the query in retrieval.
Conditioned on both $n$ and the retrieved knowledge $K_r$, LLM $M$ then generates a draft program $p^0$.
We apply the a compiler or interpreter to execute $p^0$ on LLM-generated inputs $I=[i_1, i_2, ..., i_n]$ (more details on Appendix~\ref{app:test_inputs}), which 
provides execution feedback $F^0=[f^0_1, f^0_2, ..., f^0_n]$.
Based on $n$, $p^0$, $I$, and $F^0$, we prompt an LLM $M_q$ to write a new query $q_1$ on what knowledge is currently required.
In general, given $n$, $p^i$, $I$, and $F^i$ in the iteration $i$, $M_q$ writes $q_{i+1}$, which is used for retrieval in the iteration $i+1$.

\subsection{Knowledge Soup}
\label{sec:knowledge_soup_construction}
In this section, we first introduce the four components included in the construction of the knowledge soup $K$ and then describe the process of its evolution.
\subsubsection{Construction}
We consider four types of knowledge as follows:


\paragraph{Web search} is a general and popular resource applied in traditional RAG applications.
Human programmers frequently refer to it when they try to understand some syntax or fix a bug.
It contains diverse information including blogs, tutorials, and community Q\&A discussions relevant to solving coding problems.
Intuitively, it is valuable as human programmers frequently rely on search engines to seek assistance when struggling with coding challenges.
Previous work \cite{nakano2021webgpt} has fine-tuned GPT-3 \cite{brown2020language} to answer long-form questions using a text-based web-browsing environment.
In this study, we investigate the efficacy of LLMs in utilizing web search content to solve unfamiliar coding problems without further training.
We use the Python API of Google search~\footnote{https://pypi.org/project/google/} to retrieve top-ranking websites and further convert the HTML page to markdown using the package html2text~\footnote{https://pypi.org/project/html2text/}.
In Appendix~\ref{app:web_content}, we include more discussions about the content in the web search.


\paragraph{Documentation} is commonly accessible upon the release of a new programming language or an updated library version. 
Official documentation serves to thoroughly elucidate the essential syntax and grammar required for coding. 
\citet{zhou2022docprompting} demonstrated that language models can effectively leverage code documentation after finetuning.
In this work, we focus on understanding the capability of LLMs in utilizing the documentation of updated libraries or long-tail programming languages in code generation, without making any parameter update.

\paragraph{Execution feedback} is a specific knowledge type for code generation.
It exposes syntax mistakes and locates code errors, which are frequently referenced by human programmers to debug.
While multiple types of execution can provide feedback (e.g., execution by LLMs), we focus on the compiler or interpreter execution in this work. 
Previous works~\cite{shinn2024reflexion} have demonstrated that LLMs are capable of repairing buggy code using the execution feedback.
Instead of only leveraging the error messages obtained from executing the generated faulty programs, we further enrich the knowledge base by preparing sample code-error pairs.
More details can be found in Appendix~\ref{app:sample_collect}.

\paragraph{Code snippets} are the short pieces of code that demonstrate sample usage of certain functions or syntax. 
Different from other types of knowledge that involve natural language, code snippets in programming language naturally align with the LLM generation objective and provide concrete examples of inputs, outputs, and parameters.
Furthermore, they serve as a means to convey information about the programming language itself, providing crucial details such as bracket placement, utilization of special tokens, and other grammar.
Before evaluation, we collect a set of code snippets verified to be free of syntax errors (more details in Appendix~\ref{app:sample_collect}).
Additionally, we also accumulate code solutions generated by LLMs.

\subsubsection{Evolution}
The evolution of knowledge bases is primarily contributed by the execution feedback and code snippets.
In each iteration, we execute the generated program with sample inputs (Appendix \ref{app:test_inputs}).
If the execution successfully exits, we classify the code snippet as ``syntax-correct'', which can serve as a demonstration for other instances to refer to.
Otherwise, we add the \texttt{(code snippet, error messages)} pair to the knowledge base.
Throughout the process of iterative generation, the knowledge base evolves to include increasingly rich information.

\begin{algorithm}[t]
\small
\caption{\ourmethod Pipeline}
\label{alg:evor}
\begin{algorithmic}[1]
\State \textbf{Input:} $n$: the coding problem description; $M$: the LLM to generate the code answer; $M_q$: the LLM to evolve queries; $M_t$: the LLM to generate test inputs; $R$: the retriever to output a list of relevant passages; $K$: the knowledge base; $m$: the maximum number of iterations; $E$: the compiler or interpreter to execute programs
\State \textbf{Initialization:} $I=[]$, $p=null$.
\For{$i = 0, \ldots, m$} \label{line:it-start}
    \If{$i$ = 0}
    \State $q_i \gets n$
    \Else
    \State $q_i \gets M_q(n, p^{i-1}, I, F^{i-1})$ \Comment{Evolve query}
    \EndIf
    \State $K_r \gets R(q_i,K)$   \Comment{Retrieve relevant knowledge}
    \State $p^i \gets M(n,K_r)$       \Comment{Generate program}
    \State $p \gets p^i$
    \State $F^i = E(p^i, I)$                \Comment{Execute and get feedback}  
    \If{$F^i$ is sucess}
    \State $K \gets K \cup \{p^i\}$  \Comment{Evolve knowledge base}
    \Else
    \State $K \gets K \cup \{(p^i,F^i)\}$  \Comment{Evolve knowledge base}
    \EndIf
    \If{$i$ = 0}
    \State $I \gets M_t(n,p^i)$  \Comment{Generate test inputs}
    \EndIf
    \If{terminate condition is satisfied}
    \State \textbf{break}
    \EndIf
\EndFor\label{line:it-ends}
\State \textbf{Return:} $p$: output code.
\end{algorithmic}
\end{algorithm}

\subsection{\ourmethod Pipeline}
\label{sec:active_retrieval_pipelines}
Algorithm~\ref{alg:evor} demonstrates the \ourmethod pipeline. 
In every iteration $i$, we first formulate the query $q_i$ (\S \ref{sec:query_formulation}), and use it to retrieve relevant information $K_r$ from the knowledge base $K$.
The program $p^i$ is then generated conditioned on both $n$ and $K_r$.
We get the execution feedback $F^i$ by executing $p^i$ on LLM-generated test inputs $I$ using the compiler or interpreter $E$.
The knowledge base $K$ is then evolved to include $p^i$ if the execution is successful, or the pair of ($p^i$, $F^i$) otherwise.
The pipeline exits either upon reaching the termination condition or the maximum iteration steps.
In the experiments, we set the maximum iterations to 30 and the termination condition to be the same execution feedback in consecutive 3 iterations, i.e., the algorithm exits if the program is successfully executed or results in the same error in consecutive 3 iterations.

\subsection{Datasets}
\label{sec:datasets}
Since LLMs are extensively trained on public data, we curate a new benchmark to evaluate their generalization capability with \ourmethod.
Specifically, we introduce four datasets where two focus on updated libraries and two are about long-tail programming languages.

We first modified two popular Python libraries, Scipy and Tensorflow, to simulate the real updates~\footnote{We do not use a real library update version because it is potentially exposed to LLM training data, which deviates from our purpose to evaluate LLMs' generalization ability.}, and denote them as \newscience and \jumptensor respectively. 
We then collect problems of the Scipy and Tensorflow split from DS-1000~\citep{lai2023ds} and adapt them to our modified version.
For the long-tail programming languages, we select Ring and Pony.
They have little public data and are excluded from the StarCoder training set, which involves 88 mainstream programming languages~\cite{li2023starcoder}.
We make use of the problems in LeetCode~\footnote{\url{https://leetcode.com/problemset/}} for these two datasets.
For each problem in modified libraries or long-tail programming languages, we manually write the ground truth solution and annotate the oracle documentation based on it.
We present the dataset statistics in Table \ref{tab:data_statistic}.
More details about our curation process can be found in Appendix~\ref{app:dataset_curation}.
\input{tables/data_statistics}

%% file: tables/data_statistics.tex
\begin{table}
\centering
\small
\resizebox{0.5\textwidth}{!}{%
\begin{tabular}{ccccccc}
\toprule
Dataset & \# P & \# D & A.T & A.P.L & A.S.L &  A.D.L  \\
\midrule
Scipy-M & 142 & 3920 & 3.1 & 322.6 & 44.1 & 499.7\\
Tensor-M & 45 & 5754 & 4.1 & 234.5 & 39.0 & 517.6\\
Ring & 107 & 577 & 18.2 & 108.3 & 98.3 & 334.0 \\
Pony & 113 & 583 & 18.4 & 116.9 & 129.8 & 3204.0 \\
\bottomrule
\end{tabular}
}
\caption{\label{tab:data_statistic}Data statistics of four benchmarks. We report the number of problems (\# P), the number of official documentation files (\# D), the average number of test cases (A.T), the average problem length (A.P.L), the average solution length (A.S.L) and the average gold documentation length (A.D.L). Tensor-M refers to \jumptensor. Problem length, solution length and document length are calculated by the tiktoken (https://pypi.org/project/tiktoken/) package with model gpt-3.5-turbo-1106.}
\end{table}

%% file: texts/experiment.tex
\input{tables/main1}

\section{Experiment}
\label{sec:experiment}
To verify the effectiveness of \ourmethod, we conduct extensive experiments with both the proprietary model ChatGPT (gpt-3.5-turbo-1106\footnote{https://platform.openai.com/docs/models/gpt-3-5-turbo}) and the open-source model CodeLlama~\footnote{https://huggingface.co/codellama/CodeLlama-34b-Instruct-hf}.
In \S\ref{sec:baselines}, we describe 5 baseline settings of other code generation approaches and specify the default configuration of \ourmethod in \S\ref{sec:evor_config}.
In \S\ref{sec:result}, we compare the results of \ourmethod, existing code generation methods, as well as their combinations.
By default, we use the execution accuracy (pass@1) as the metric throughout the paper.

\subsection{Baselines}
\label{sec:baselines}
We compare \ourmethod to the vanilla generation and four recent methods that demonstrate significant performance improvement in code generation tasks:

\textbf{Vanilla}: we implement the vanilla generation baseline where we directly get the outputs from LLMs based on the coding question $n$ without augmenting external knowledge.

\textbf{MPSC}: \citet{huang2023enhancing} proposed Multi-Perspective Self-Consistency (MPSC) incorporating both inter- and intra consistency.
Following the original implementation, we prompt LLMs to generate diverse outputs from three perspectives: Solution, Specification and Test case,  construct the 3-partite graph, and pick the optimal choice of solutions based on confidence scores.

\textbf{ExeDec}: \citet{shi2023exedec} introduced a decomposition-based synthesis strategy, where they employ a subgoal model to predict the subgoal of the desired program state for the next part of the program and use another synthesizer model to generate the corresponding subprogram to achieve that subgoal.
Subprograms are finally combined as the output answer to solve the original coding problem.
In experiments, we use ChatGPT as the subgoal model and compare LLMs to synthesize programs following the subgoal predictions.

\textbf{Reflexion}: \citet{shinn2024reflexion} uses a framework to reinforce LLMs through linguistic feedback.
It employs an iterative optimization process.
In each iteration, the actor model produces a trajectory conditioned on the instructions and memories.
The evaluator model then evaluates the trajectory and calculates a scalar reward.
Self-reflection model generates verbal experience feedback on the pairs of trajectories and rewards, which are stored in the memory.
Throughout experiments, we use the compiler or interpreter as the evaluator model, which returns 0 upon execution errors, and 1 otherwise.
By default, we use ChatGPT as the self-reflection model and compare the capabilities of LLMs to generate programs as actor models.

\textbf{DocPrompting}: \citet{zhou2022docprompting} proposed to explicitly leverage code documentation by first retrieving the relevant documentation pieces given a natural language (NL) intent, and then generating code based on the NL intent and the retrieved documentation. 
It can be viewed as a degraded version of \ourmethod where neither queries nor knowledge bases evolve and the retrieval pool encompasses the documentation as a single source.

\subsection{Default \ourmethod Configuration}
\label{sec:evor_config}
By default, we incorporate the documentation, execution feedback, and code snippets in the knowledge soup $K$ for \ourmethod, as the content of web search contains large portions of noisy information (Appendix~\ref{app:web_content}) and only marginally improves the results (\S\ref{sec:knowledge}).
We use ChatGPT for both $M_q$ to evolve queries and $M_t$ to generate test inputs, and vary $M$ between ChatGPT and CodeLlama to output code answers.
We employ the INSTRUCTOR-xl~\cite{su-etal-2023-one} as the primary retrieval model~(Appendix~\ref{app:retriever}) and allow a maximum context length of 4,096 for both ChatGPT and CodeLlama, as the further increase incurs a higher cost, but fails to provide additional improvements (Appendix~\ref{app:long_context_model}).

\subsection{Results}
\label{sec:result}
Table~\ref{tab:main} shows that existing code generation approaches perform poorly on \ourdata.
With CodeLlama, the improvements of MPSC, ExeDec, and Reflexion are smaller than 2\% on average, compared to the vanilla generation.
In particular, the execution accuracy remains 0 in Ring across three methods.
This indicates that, even though existing approaches excel in code generation tasks that do not require external knowledge (e.g., HumanEval~\cite{chen2021evaluating}), they cannot be directly applied to the setting of RACG without designing extra mechanisms to retrieve and utilize the external information.
In contrast, by explicitly using documentation, DocPrompting significantly surpasses MPSC, ExeDec, and Reflexion by a large margin, further confirming that domain knowledge is critical to solving tasks in \ourdata.

Furthermore, \ourmethod achieves 16.1\% and 16.2\% absolute gain with ChatGPT and CodeLlama respectively on top of DocPrompting.
This can be explained by the fact that DocPrompting only uses the documentation as a single retrieval source, without evolution in both queries and knowledge.
By combining \ourmethod with MPSC, ExeDec, or Reflexion, we observe further performance increase by up to 2.6\% on average with ChatGPT.
This suggests that \ourmethod is flexible to be integrated with existing approaches to further push forward the boundary of LLM performance in RACG.




%% file: tables/main1.tex
\begin{table*}
\centering
\small
\begin{tabular}{lc ccccc c cccc}
\toprule
\textbf{Method} & \multicolumn{5}{c}{\textbf{Model: ChatGPT}} & & 
\multicolumn{5}{c}{\textbf{Model: CodeLlama}}\\
\cmidrule{2-6}
\cmidrule{8-12}
 & \newscience & Tensor-M & Ring & Pony & Avg. & & \newscience & Tensor-M & Ring & Pony  & Avg. \\
\midrule
 & \multicolumn{11}{c}{\textit{Baseline}} \\
Vanilla & 17.6 & 11.1 & 3.7 &  1.8 & 8.6 & & 11.3 & 17.8 & 0.0 & 0.0 & 7.3\\
MPSC & 18.3 & 11.1 & 4.1 & 1.8 & 8.8 & & 11.6 & 17.8 & 0.0 & 0.0 & 7.4\\
ExeDec & 22.5 & 17.8 & 4.5 & 3.6 & 12.1 &  & 13.2 & 17.8 & 0.0 &  0.0 & 7.8\\
Reflexion & 23.2 & 22.2 & 5.3 & 4.7 & 13.9 &  & 14.5 & 20.0 & 0.0 &  0.9 & 8.9\\
DocPrompting & 32.4 & 33.3 & 8.4 & 2.7 & 19.2 &  & 16.9 & 37.8 & 4.7 & 4.4 & 16.0 \\
\midrule
& \multicolumn{11}{c}{\textit{Ours}} \\
\ourmethod & 37.9 & 53.3 & 36.6 & 13.5 & 35.3 &  & 31.2 & 53.3 & 26.7 & 17.4 & 32.2\\
\ourmethod + MPSC & 38.6 & 55.6 & 37.8 & 15.6 & 36.9 &  & 33.6 & 55.6 & 27.3 & 18.4 & 33.7 \\
\ourmethod + ExeDec & 39.2 & 55.6 & 40.0 & 16.3 & 37.8 &  & 34.1 & 57.8 & 27.9 & 19.1 & 34.7 \\
\ourmethod + Reflexion & 39.4 & 55.6 & 39.2 & 17.3 & 37.9 & & 35.3 & 55.6 & 28.6 & 18.8 & 34.6\\
\bottomrule
\end{tabular}

\caption{The performance of baseline methods, \ourmethod and their combinations in \ourdata. \ourmethod demonstrates significantly superior results, with further improvement when combined with other baseline methods. \label{tab:main}
}
\end{table*}

%% file: texts/analysis.tex
\section{Analysis}
\input{tables/synchronous}
\subsection{Synchronous evolution}
\label{sec:syn}
We investigate how the synchronous evolution of queries and knowledge influences the RACG performance of LLMs.
We compare \ourmethod to the setting where we only evolve queries (skip line 13-20 in Algorithm~\ref{alg:evor}), only evolve knowledge (skip line 7 in Algorithm~\ref{alg:evor}), and evolve neither of them (skip line 7, 13-20 in  Algorithm~\ref{alg:evor}, and terminate in a single iteration).
We adopt the default setting in \S\ref{sec:evor_config} except the specified changed in the algorithm.
Table~\ref{tab:synchronous} shows evolving either queries or knowledge significantly enhances the results, highlighting that knowledge evolution also contributes to improving RACG in addition to the query rewriting.
By applying the synchronous evolution of both queries and knowledge, \ourmethod consistently outperforms evolving either of them by large margins across all datasets in \ourdata.
This suggests the complementary strength of synchronous evolution for eliciting the best performance of LLMs in RACG.

\input{tables/knowledge1}

\subsection{Diverse Knowledge}
\label{sec:knowledge}
To understand the influence of diverse knowledge sources on \ourmethod, we conduct an ablation study by constraining the types of knowledge in the retrieval pool.
Specifically, we construct the knowledge soup $K$ with only one of web search, execution feedback, code snippets and documentation.
We also consider the pairwise combination of execution feedback, code snippets and documentation, and the setting where all of them are included.
For each constructed knowledge soup, we experiment with evolving neither queries nor knowledge, and evolving both of them.
We skip line 14 in Algorithm~\ref{alg:evor} when the code snippets are not included in the knowledge soup and skip line 16 when the execution feedback is not incorporated.
Exceptions occur when the knowledge soup consists solely of web search content or documentation, where we only evolve queries.

In Table~\ref{table:knowledge}, we present the average performance of ChatGPT and CodeLlama using different types of knowledge sources, under two settings where evolution is and is not involved.
The results show that, when augmenting with code snippets or syntax documentation, the performance of ChatGPT and CodeLlama is significantly higher than those using the web search or execution feedback.
In particular, both models achieve less than 1\% improvement when only using the web search as the knowledge source without evolving queries. 
This indicates that the general web search may not provide the most effective information to adapt LLMs in RACG.
Compared to single-source retrieval, LLMs consistently achieve better results when more types of knowledge are integrated.
For example, without queries and knowledge evolution, ChatGPT archives 6.2\% higher average performance by using both code snippets and documentation as the knowledge sources, compared to only employing the code snippets.
This indicates the advantage of diverse knowledge soup in enhancing the RACG performance of LLMs.

On the other hand, by evolving both queries and knowledge or only evolving queries, both ChatGPT and CodeLlama achieve significantly larger improvements when the knowledge soup becomes more diverse.
For example, when the documentation is further included in the knowledge soup on top of the execution feedback and the code snippets, CodeLlama enhances the average result from 15.9\% to 20.4\% (+4.5\%) in the setting where queries and knowledge are not evolved, but enhances from 25.3\% to 32.2\% (+6.9\%) when both are evolved.
This suggests that synchronous evolution is critical to fully exploit the advantage of diverse knowledge soup in adapting LLMs in RACG.

\subsection{Repo-level Code Generation}
\label{sec:repo}
Apart from updated libraries and long-tail programming languages, repo-level code generation is also a natural and realistic scenario for RACG, where LLMs are instructed to solve issues with reference to the Github repository code.
Different from the documentation in \ourdata, the repository code could be much more complex with intertwined variable dependencies, customized function calls, etc.
To solve an issue, the LLM usually needs to act as an agent to explore directories, use tools, make decisions, and more.
Recent efforts have demonstrated the success of such agent-based methods~\cite{opendevin2024,yang2024swe}.

\begin{figure}[t]
\begin{center}
\centerline{\includegraphics[width=\columnwidth]{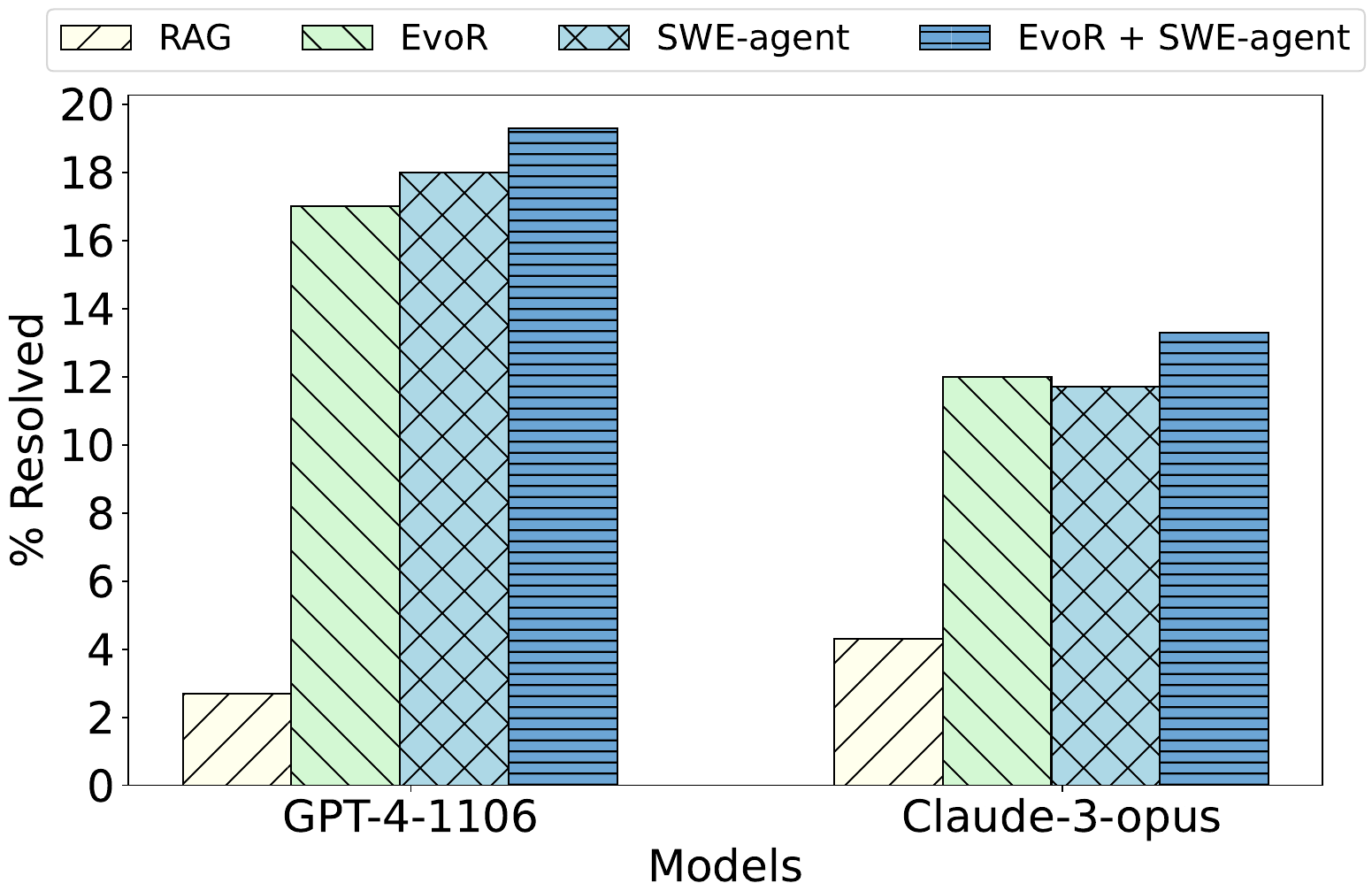}}
\caption{Performance of RAG, \ourmethod, SWE-agent (Agent) and combination of \ourmethod and SWE-agent.}
\label{fig:swe}
\end{center}
\vskip -0.2in
\end{figure}

We explore the applicability of \ourmethod in this challenging setting.
Specifically, we employ the popular SWE-bench-Lite~\cite{jimenez2023swe} as the testbed, use all the repository content as the documentation, and adopt the configuration in \S\ref{sec:evor_config}.
Due to the difficulty of the tasks, we experiment with two settings: (1) use GPT-4-1106 for all LLMs in Algorithm~\ref{alg:evor}; (2) use Claude-3-opus.
Figure~\ref{fig:swe} shows that \ourmethod outperforms the traditional RAG by a large margin, and is comparable with SWE-agent.
This highlights the generalizability of \ourmethod with successful application in repo-level code generation.

Furthermore, we integrate \ourmethod with SWE-agent where we augment the search space of SWE-agent to include the execution feedback and code snippets without syntax errors, and dynamically update queries and the knowledge base in every iteration of generation.
Figure~\ref{fig:swe} demonstrates additional performance improvements on top of both \ourmethod and SWE-agent, further proving \ourmethod's flexibility in its integration to agent-based approaches.

\subsection{Effective Token Usage}
\label{sec:token}
\begin{figure}[t]
\begin{center}
\centerline{\includegraphics[width=\columnwidth]{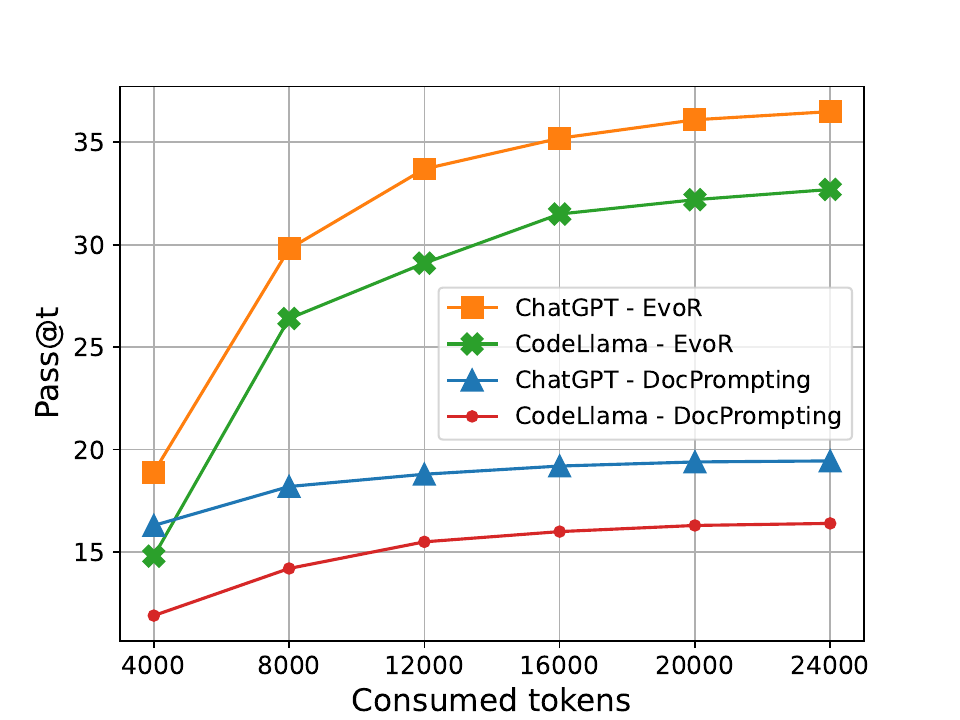}}
\caption{The pass rate of ChatGPT and CodeLlama at different token consumption levels. The results show that \ourmethod achieves a more significant increase compared to DocPrompting when the consumed tokens increase.}
\label{fig:efficiency}
\end{center}
\vskip -0.2in
\end{figure}

\citet{olausson2023self} argues that iterative code generation, e.g., self-repair, may not yield higher pass rates when taking the cost into account.
We conduct additional experiments to check the performance of \ourmethod at different levels of token budgets.
In Algorithm~\ref{alg:evor}, we adopt the termination condition as the limit of maximum token consumption, i.e., the algorithm exits when the tokens used by LLMs throughout iterations exceed a given threshold.
Additionally, we compare \ourmethod to DocPrompting, the best baseline approach in Table~\ref{tab:main}.
Following~\citet{olausson2023self}, we sample LLMs multiple times until the token limit, using the concatenation of the given question $n$ and the retrieved documentation $K_r$ as the prompt.
We calculate pass@t and set the token threshold to 4,000, 8,000, 12,000, 16,000, 20,000 and 24,000.
Figure~\ref{fig:efficiency} shows that \ourmethod achieves significantly higher performance at all token levels for both ChatGPT and CodeLlama.
With the increase of consumed tokens, \ourmethod demonstrates larger improvements compared to DocPromting, indicating the more effective token usage of \ourmethod in generalizing LLMs in RACG.

%% file: tables/synchronous.tex
\begin{table}[t]
\centering
\small
\scalebox{0.85}{
\begin{tabular}{ccccccc}
\toprule
evolution & \newscience & Tensor-M & Ring & Pony & Avg   \\
\midrule
& \multicolumn{5}{c}{\textit{Model: ChatGPT}} \\
 No evolution & 32.6 & 40.0 & 11.7 & 5.2 & 22.4\\
 Evolve query & 32.9 & 44.4 & 27.8 & 8.5 & 28.4 \\
 Evolve knowledge  & 33.5 & 42.2 & 13.5 & 6.1 & 23.8\\
 \ourmethod (Evolve both) & 37.9 & 53.3 & 36.6  & 13.5 & 35.3 \\
\midrule
& \multicolumn{5}{c}{\textit{Model: CodeLlama}} \\
 No evolution & 23.9 & 42.2 & 8.2 & 7.3 & 20.4\\
 Evolve query & 26.6 & 44.4 & 11.7 & 12.8 & 23.9 \\
 Evolve knowledge  & 25.8 & 44.4 & 12.6 & 8.3 & 22.8 \\
 \ourmethod (Evolve both) & 31.2 & 53.3 & 26.7 & 17.4 & 32.2 \\
\bottomrule
\end{tabular}
}
\caption{The performance of ChatGPT and CodeLlama when neither queries nor knowledge evolves, only the query evolves, only the knowledge evolves and when both evolve (\ourmethod). Results show that evolving both is consistently better across all datasets.\label{tab:synchronous}
}
\end{table}

%% file: tables/knowledge1.tex
\begin{table}
\centering
\small
\resizebox{0.5\textwidth}{!}{
\begin{tabular}{lcc c cc}
\toprule
evolution ($\rightarrow$) & \multicolumn{2}{c}{w/o evolution} & & \multicolumn{2}{c}{w/ evolution} \\
\cmidrule{2-3}
\cmidrule{5-6}
Knowledge ($\downarrow$) & ChatGPT  & CodeLlama & & ChatGPT  & CodeLlama\\
\midrule
None & 8.6 & 7.3 & & - & -\\
\midrule
& \multicolumn{5}{c}{\textit{Single Source Retrieval}} \\
Web & 9.7 & 7.5 & & 10.6 & 8.2 \\
Exec & 11.7 & 7.4 & & 13.8 & 9.1\\
Code & 15.6 & 16.2 & & 23.5 & 24.8\\
Doc & 19.2 & 16.0 & & 28.9 & 20.5 \\
\midrule
& \multicolumn{5}{c}{\textit{Knowledge Soup Retrieval}} \\
Exec + Code & 18.3 & 15.9 & & 27.8 & 25.3 \\
Exec + Doc & 20.7 & 17.2 & & 32.4 & 23.0\\
Code + Doc & 21.8 & 19.6 & & 33.5 & 31.2\\
Exec + Code + Doc & 22.4 & 20.4 & & 35.3 & 32.2\\
\bottomrule
\end{tabular}
}
\caption{The average performance of ChatGPT and CodeLlama with different knowledge sources. Web refers to web search, Exec refers to execution feedback, Code refers to code snippets and Doc refers to documentation. The results show that diverse types of knowledge enhance RACG performance, where the improvement is larger under the setting with evolution.\label{table:knowledge}
}
\end{table}

%% file: texts/related_works.tex
\section{Related works}

Since the focus of our work is to enhance code generation with retrieval, our work is closely related to code generation and retrieval-augmented code generation. Additionally, we are connected to the line of code execution since we also leverage it as an important retrieval source.

\paragraph{LLM-based Code Generation}

LLMs that have been pre-trained on extensive code corpus have exhibited impressive abilities in the domain of code generation \cite{li2022competition,nijkamp2022codegen,li2023starcoder,roziere2023code,wei2023magicoder}.
Numerous techniques have been suggested to improve the coding capabilities of LLM without the need to adjust its parameters \cite{chen2022codet,huang2023enhancing,li2023chain,zhang2023algo,chen2023teaching,key2022speak}
However, most of these works set up the evaluation in scenarios LLMs are familiar with, e.g., HumanEval \cite{chen2021evaluating}, HumanEvalPack~\citep{octopack2024} and MBPP \cite{austin2021program}, where they are capable of demonstrating superior zero-shot performance by only utilizing internal knowledge.
In this work, we focus on evaluating the capabilities of LLMs to incorporate external knowledge for the purpose of code generation in updated libraries or less-common programming languages. Our task reflects a more realistic yet challenging scenario for LLMs.

\paragraph{Retrieval-Augmented Generation}

The retrieval-augmented generation (RAG) is an appealing paradigm that allows LLMs to efficiently utilize external knowledge~\cite{shi2023replug,izacard2020leveraging,xu2023recomp,jiang-etal-2023-active}.
Recent works have applied it to the code generation task~\cite{patel2023evaluating, guo2023retrieval, parvez-etal-2021-retrieval-augmented,wang2023rap}.
Specifically, \citet{zhou2022docprompting} explored the natural-language-to-code generation approach that explicitly leverages code documentation.
\citet{zan2022language} introduced a framework designed to adapt LLMs to private libraries, which first utilizes an APIRetriever to find useful APIs and then leverages an APICoder to generate code using these API docs.
\citet{zhang-etal-2023-repocoder} employed the iterative generate-retrieval procedure to do repository-level code completion.
There are also recent efforts showing much enhanced performance by simply rewriting the queries~\cite{ma2023query,anand2023context,chan2024rq}
To the best of our knowledge, we are the first to adopt the synchronous evolution of queries and diverse knowledge to explore the setting where LLMs need to incorporate external information in code generation.

\paragraph{Code Execution}

Previous works have extensively employed executors (Interpreters/Compilers) in code-related tasks~\cite{wang2022execution,liu2023code,olausson2023demystifying,chen2023teaching}.
\citet{shi2022natural} introduced execution result–
based minimum Bayes risk decoding for program selection.
\citet{yang2023intercode} established an interactive coding benchmark by framing the code as actions and execution feedback as observations.
\citet{chen2023teaching} use the execution result as feedback to help LLM refine the code.
In this work, we utilize the executor to provide feedback and check code outputs on syntax errors, which contributes to evolve both queries and knowledge in RACG.

%% file: texts/conclusion.tex
\section{Conclusion}
Much recent work illustrated the ability of LLMs to incorporate external knowledge with retrieval-augmented generation.
We propose a novel pipeline, \ourmethod, which achieves two to four times execution accuracy compared to existing code generation methods.
Extensive experiments demonstrate that \ourmethod can be easily combined with them to provide further improvements, including the agent-based ones to solve challenging tasks such as repo-level code generation.
Through an in-depth analysis, we further show the complementary strength of synchronous evolution of queries and documents in RACG, which enhances the model performance by larger margins with more diverse knowledge sources.
We hope that our findings will inspire researchers and practitioners to develop efficient and effective strategies in their customized code-generation tasks with LLMs.

%% file: texts/limitation.tex
\section{Limitations}
Despite the effectiveness of \ourmethod in RACG, one limitation is that it requires multiple rounds of interactions among retrievers, LLMs, and executors to output the code answer. 
This iterative process can lead to longer latency and increased energy consumption, which are critical concerns in real-time applications and energy-constrained environments. 
We hope that future work will design more efficient architectures or approaches to integrate LLMs seamlessly while maintaining or improving performance in RACG. 
Such advancements could significantly enhance the practicality and scalability of LLMs in realistic scenarios.

\section{Potential Risk}
The use of retrieval-augmented code generation with large language models introduces several potential risks, primarily centered around the quality and relevance of the retrieved code snippets. 
There is a risk of biased or incorrect information being retrieved, which could propagate errors or introduce vulnerabilities into generated code. 
Additionally, there are concerns about privacy and security if sensitive code snippets are inadvertently included in the retrieval process. 
Addressing these risks requires careful curation of retrieval sources, robust validation mechanisms, and continuous monitoring to ensure the integrity and safety of the generated code.

%% file: texts/appendix.tex
\appendix
\section{Dataset curation}
\label{app:dataset_curation}
We introduce more details about our dataset curation process for updated library (\S\ref{app:updated_library_data}) and long-tail programming languages (\S\ref{app:long_tail_language_data}).
In \S\ref{app:testcase_generation}, we describe our implementation of the test case construction for the dataset Ring and Pony.

\subsection{Library-oriented data collection}
\label{app:updated_library_data}
Following \citet{zan2022language}, we use the synonyms of original API names and API arguments in the updated library, such as converting \texttt{stack} to \texttt{pile}.
Additionally, we combine two similar APIs into one, with newly added arguments to distinguish the authentic functionalities, e.g.,  \texttt{linear\_interpoloate} integrates two APIs, \texttt{griddata} and \texttt{interp1d}. 
Finally, we create new class objects and align methods with the original class. 
For instance, a new \texttt{SparseMatrix} object is created to include all sparse matrix objects in Scipy. 
We rewrite the ground truth solution for each example with new APIs.

To construct the documentation of the updated libraries, we first collect the original libraries~\footnote{\url{https://docs.scipy.org/doc/}, \url{https://www.tensorflow.org/api_docs}}. 
We then replace the old documentation files with our modified version. 
For each question, we annotate the oracle documentation by checking the ground truth answer. 
We grasp the corresponding documentation pages and concatenate them to serve as the minimum documentation required for answering the problem.
We reuse the test cases introduced in DS-1000 to evaluate LLM generalization performance.

\subsection{Language-oriented data collection}
\label{app:long_tail_language_data}
For each programming problem collected from LeetCode, we rewrite the function signatures to adapt them to the target programming language.
We collect the whole documentation for Ring and Pony from their websites: \url{https://ring-lang.github.io/doc1.19/} and \url{https://www.ponylang.io/}. 
For each question, we labeled the oracle documentation of the specific grammar used in the ground truth, such as data structures or branching syntax. 
We concatenate the document for each new syntax used in the ground truth to obtain a minimum document that contains the required syntaxes for answering the question.


\subsection{Test-case generation for language-oriented data}
\label{app:testcase_generation}
To accurately evaluate the performance of LLM in writing code of long-tail programming languages, we follow \cite{liu2023your} to construct a comprehensive set of test cases for each problem.
Specifically, we first prompt ChatGPT to write a validation script and solution script using Python. 
The validation script will check for the input constraints (e.g. single line of a positive integer, two binary strings, etc.) of the problem. 
The solution script is supposed to generate the correct answer given a valid input.   
We then manually check both scripts and modify them if necessary for all problems. 
Next, we prompt ChatGPT to create complex and corner cases until there are 20 test cases for each problem. 
We then apply mutation-based strategies to extend the number of test cases for each problem to 200. The mutation-based strategy works as follows. We first parse the input into the appropriate format and types (e.g. list of strings, tuple of integers, etc. ) We will then randomly mutate the test cases multiple times to create a new input based on the types. For instance, we may add 1 or subtract 1 from an integer to mutate it.
All generated test cases added are checked by both the validation script and solution script.
A test case is considered as valid if the following three conditions are met: (1). Both scripts do not report any error; (2). The solution script terminates within 1 second; (3). The answer returned by the solution script matches that in the test case.

The final step is to apply test-suite reduction which selects a subset of all input test cases while preserving the original test effectiveness (i.e. the reduced set of test cases marks a code solution as right/wrong if and only if the original set marks it as right/wrong). 
We employ the three strategies proposed by \cite{liu2023your}: code coverage, mutant killing, LLM sample killing. Code coverage evaluates how each test case covers different branch conditions in the solution script. Mutant killing employees a mutation testing tool for Python to create mutant codes from the solution script. LLM sample killing prompts llama-2-70b to generate several incorrect solutions to the problem. We run all test cases against these different codes to perform the test-suite reduction.
Finally, we generate the answer using the solution scripts.


\section{Prvate Library}
We notice that \citet{zan2022language} crafted three benchmarks named TorchDataEval, MonkeyEval, and BeatNumEval to evaluate the capability of language models in code generation with private libraries.
Their benchmarks share some similarities with our two datasets on updated libraries, where we both modified popular Python libraries to explore the setting for LLM generalization.
Different from them, our datasets are built with increased complexity, where we not only use the simple synonym to update the API names, but additionally combine two APIs and create new class objects.
This indicates that our datasets are likely to cover broader scenarios of library updates in real life.

Nonetheless, we also benchmark our system on their datasets with varied knowledge source.
Table \ref{table:private_lib} shows that CodeLlama achieves exceptionally high score in all three datasets, with zero-shot accuracy 80.2\% in Monkey.
Since the three datasets were available in Github as early as 2022, which is well ahead of the time CodeLlama was released, we suspect that CodeLlama has been trained on the three datasets.
Although our system still looks to be effective in their benchmarks with performance gain by including more knowledge sources, we are concerned that these datasets may not be able to reflect the generalization capabilities of LLM.


\input{tables/private_library}

\section{LLM-generated program inputs}
\label{app:test_inputs}
\input{tables/test_inputs}
To verify the syntax of the generated program, one effective way is to execute it with test cases.
To simulate the scenario where no test case is available, we investigate whether it is possible to generate program inputs with LLMs.
Specifically, we prompt ChatGPT and CodeLlama to generate 5 test cases for each problem, and only save the inputs for evaluating the syntax of other programs.
As an ablation study, we execute the gold program of each problem with the generated inputs and count a generated input as valid if no error is reported during execution.
We calculate the accuracy as the percentage of examples where all the generated test inputs are valid.
Table \ref{table:test_inputs} shows that both ChatGPT and CodeLlama exhibit superior performance in generating test inputs.
This indicates that LLM-generated test inputs serve as good resources as syntax verifiers.

\section{Sample code snippets and execution feedback}
\label{app:sample_collect}
We collect sample code snippets and execution feedback in constructing the knowledge base.
Specifically, we prompt LLMs to write short scripts of sample usage of each function in the documentation corpus.
We then execute those scripts.
If the execution of a code snippet reports errors, we include it as a pair of (code, error); otherwise, we regard it as a code snippet that could demonstrate the syntax and function usage.

\section{Cost Analysis}
\input{tables/cost}
\label{app:cost_analysis}
Despite significant enhancement of \ourmethod in the generalization results, the iterative process that involves multiple LLM generations incurs large costs.
In this section, we discuss the trade-off between the cost and the performance.
To measure the cost, we count the total tokens processed by LLM throughout the process in each example.

From Table \ref{tab:cost}, we can see that, the exceptional performance is linked to the extensive processing of tokens.
Compared to employing Single-time-Q, which simulates the traditional RAG pipeline and directly uses the question as the query to retrieve documentation, ChatGPT and CodeLlama achieve 2.9\% and 4.1\% performance gain in average execution accuracy by using Single-time, which formulates the query as the explained code and retrieves from diverse knowledge soup.
This enhancement is at the expense of around 25\% more processed tokens for both models.
With active retrieval, the average performance further increases by 15.4\% and 15.1\% for ChatGPT and CodeLlama respectively.
However, the processed tokens increase by more than 2 times for both models.
With a notable increase in both cost and performance, there arises a trade-off for practitioners to carefully weigh and adjust according to their specific requirements.

\section{More Experimental Setting}
\label{app:more_exp_setting}
In all settings, we leave a length of 400 for generation and adopt ChatGPT as the LLM to explain the code, i.e., all the code is fairly translated into the explained code.
In every iteration of active retrieval and LLM generation, we add the examples with correct syntax (judged by executors with sample inputs) to the set of code snippets and only rectify the code with syntax error.

For each of the knowledge sources considered in this paper, we adopt the following principles if it is included in the prompt for LLM generation: 
(1). For web search content, include it until the maximum allowed length, e.g., 4,096, as we do not merge it without other knowledge sources;
(2). For execution feedback, include the error message and the line of the code that leads to the error;
(3). For code snippets, allocate a maximum length of 300 to them, as they are usually short;
(4). For documentation, always include other types of knowledge first, and include documentation to fill in the rest length.
For example, if we want to include both documentation and code snippets as the knowledge source and the maximum context length is 4,096, we will allocate a maximum length of 300 to code snippets and a maximum length of 4,096-300-400=3,396 to the documentation.

\section{Web Search}
\label{app:web_content}
As the artificially modified libraries are not available online, we replace the documentation returned by web search with our modified version.
In addition, we heuristically update the content from web search based on our modifications, e.g., map keywords to the synonyms we use.

\begin{figure}[ht]
\begin{center}
\centerline{\includegraphics[width=\columnwidth]{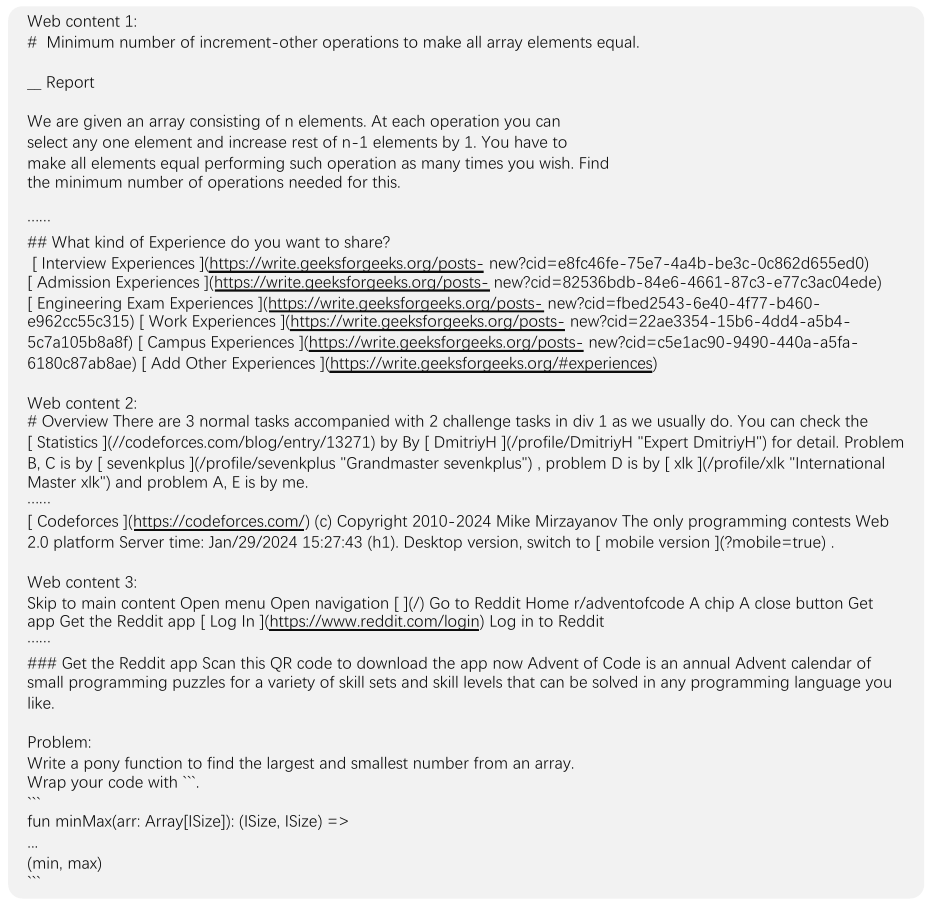}}
\caption{\label{fig:web_content_example}A web content example covering tutorials (web content 1), blogs (web content 2) and Q\&A discussions (web content 3). 
We show the top-3 results returned by google search and cut each webpage for brevity.}
\end{center}
\vskip -0.4in
\end{figure}

In Figure \ref{fig:web_content_example}, we present an example of the top-3 web search results returned by Google search to the query "In the programming language Pony, checks if the current element is less than the previous element in the nums array".
Due to the infrequent usage of the programming language Pony, there is little available resource online.
The web search fails to identify the relevant knowledge piece.
Even the specific instruction "programming language Pony" is given in the query, a guidance to solve the problem in C++ is included.
In addition, the returned texts are long, complex, and diverse, mixing various types of knowledge sources including tutorials, blogs, and community Q\&A discussions.
LLMs may find it challenging to process and effectively utilize all of the information simultaneously.
Finally, although we empirically remove some unrelated information, e.g., remove the line that starts with $*$ that is likely to be an irrelevant item listing, there is more that is hard to remove with just heuristics.
This poses a great challenge to LLMs as they are burdened to filter the unrelated content and avoid getting distracted by it.

\begin{figure}[tb]
\begin{center}
\centerline{\includegraphics[width=\columnwidth]{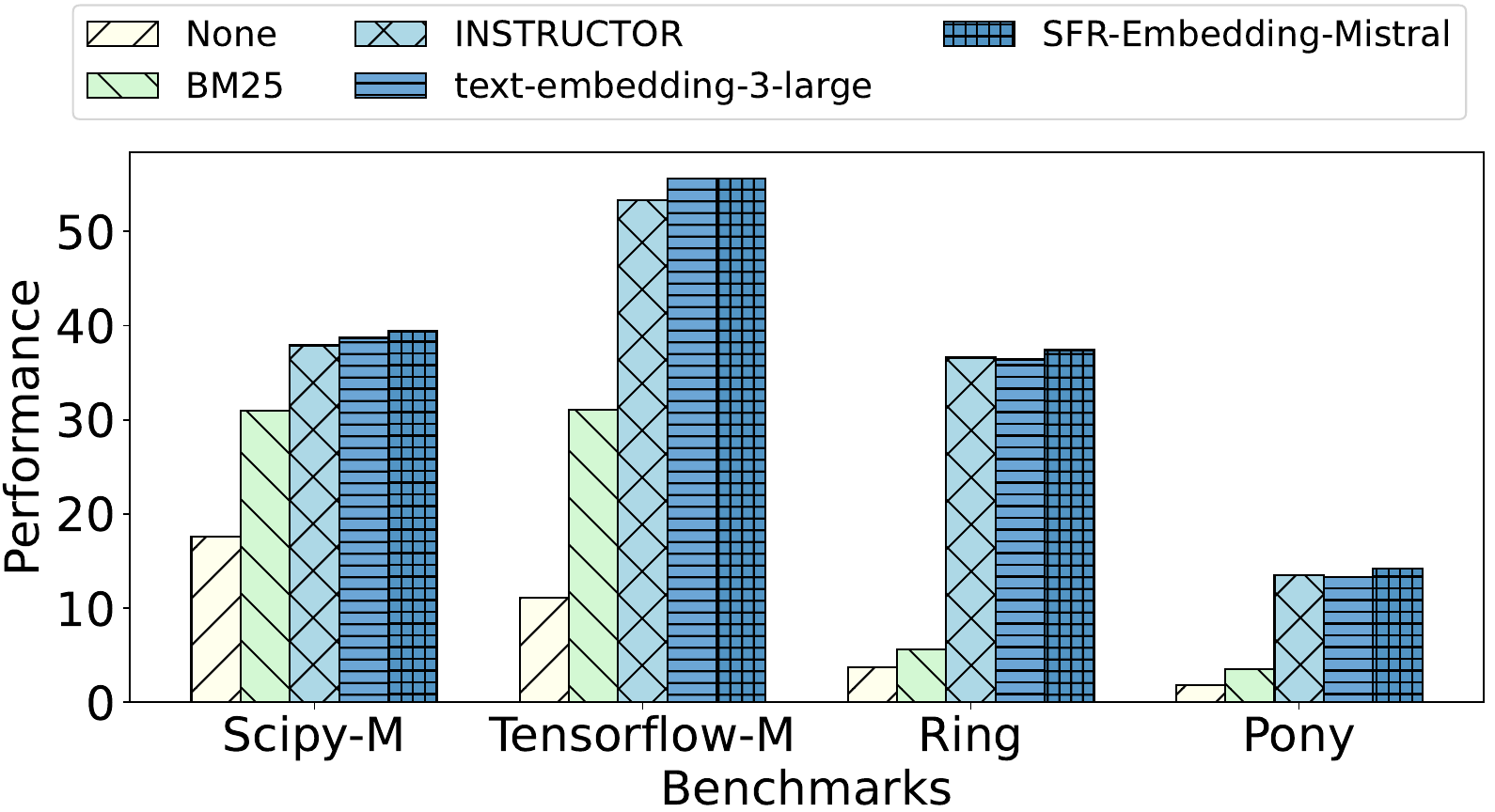}}
\caption{Comparison of ChatGPT generalization performance when the sparse retriever (BM25), or the dense retriever (INSTRUCTOR, text-embedding-3-large, SFR-Embedding-Mistral) is employed. 
The results show that dense retrievers significantly outperform their sparse counterpart, BM25.
In general, ChatGPT achieves the best performance when SFR-Embedding-Mistral is used as the retrieval model.
}
\label{fig:retrieval_model}
\end{center}
\vskip -0.2in
\end{figure}

\section{Retrieval Model}
\label{app:retriever}
We experiment with a representative sparse retriever, BM25, and several competitive dense retrievers using the default setting of \ourmethod, except for changing the retrieval model. 
INSTRUCTOR-xl~\cite{su-etal-2023-one} is an 1.3B embedding model fine-tuned to follow instructions for efficient adaptation.
text-embedding-3-large\footnote{https://platform.openai.com/docs/guides/embeddings} is OpenAI's latest embedding model, showcasing competitive performance.
SFR-Embedding-Mistral is trained on top of E5-mistral-7b-instruct~\cite{wang2023improving} and Mistral-7B-v0.1~\cite{jiang2023mistral} and achieves state-of-the-art performance in MTEB leaderboard~\cite{muennighoff2022mteb}.


As shown in Figure \ref{fig:retrieval_model}, across four datasets, when utilizing dense retrievers, ChatGPT significantly enhances the performance achieved with a sparse retriever.
Aligned with the results in the retrieval benchmark (MTEB), ChatGPT consistently achieves the best performance when using SFR-Embedding-Mistral as the retrieval model.
However, the gap between different dense retrievers is not significant.
After considering both the performance and the cost, we opt for INSTRUCTOR-xl for efficient and cost-effective development of \ourmethod.

\begin{figure}[t]
\begin{center}
\centerline{\includegraphics[width=\columnwidth]{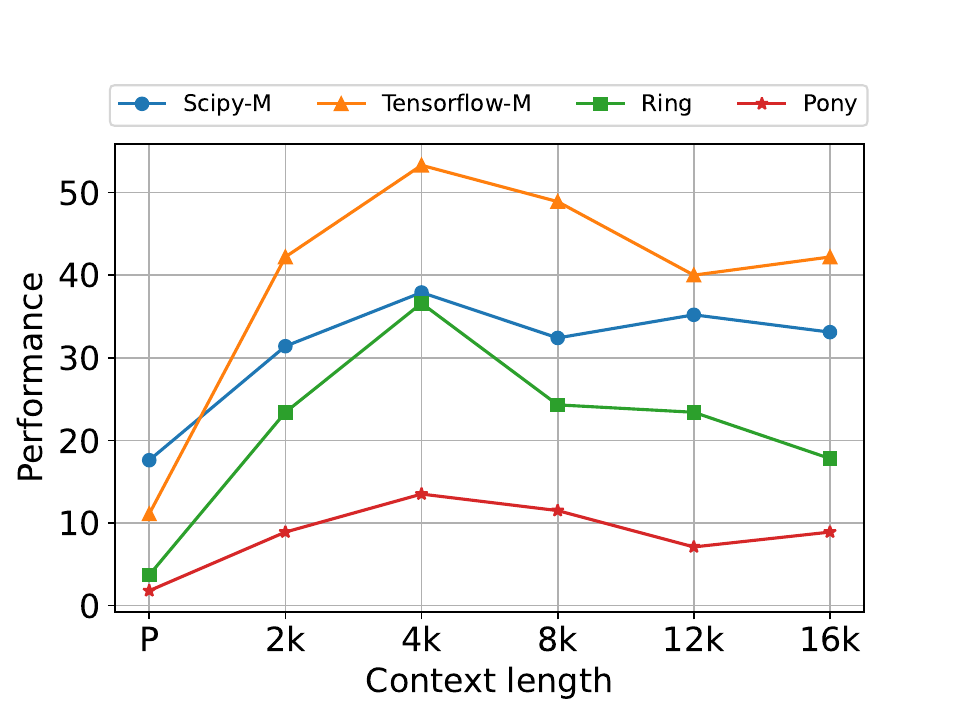}}
\caption{ChatGPT performance with various maximum allowed context lengths. 
P refers to the baseline where no external knowledge is included.
Although the model supports the context length up to 16k, the results reveal that the execution accuracy ceases to enhance when the context window is expanded from 4k to 16k.
This suggests that augmenting ChatGPT with external knowledge beyond the 4k context does not yield further improvement in the generalization performance.
}
\label{fig:long_context}
\end{center}
\vskip -0.2in
\end{figure}

\section{Long-context Model}
\label{app:long_context_model}
Besides the retrieval-based pipelines, long-context models are another alternative for LLMs to incorporate massive external knowledge.
The context window of Claude 2.1\footnote{https://www.anthropic.com/news/claude-2-1} and GPT-4\footnote{https://platform.openai.com/docs/models/gpt-4-and-gpt-4-turbo} have reached 200k and 128k tokens respectively, which questions the necessity to adopt RACG, where only a small portion of knowledge is retrieved and exposed to LLMs.
Intuitively, LLMs benefit from larger context windows, as they can utilize more external knowledge to enhance their coding.
However, our experiments do not imply the case.

We adopt the default setting of \ourmethod, but only change the maximum context length of LLMs to 2k, 4k, 8k, 12k, and 16k tokens.
Figure \ref{fig:long_context} indicates that ChatGPT achieves the best performance when only using external knowledge of 4k tokens.
This aligns with the findings in \citet{xu2023retrieval}.
With extended context lengths, i.e., more retrieved content is included in the prompt, the performance does not further increase.

The potential reasons to explain this situation include: 
(1). Only a few documents are required to answer a specific question. 
As shown in Table \ref{tab:data_statistic}, the length of gold documentation, i.e., minimum required syntax descriptions, never surpasses 4k, which does not even surpass 1k in \newscience, \jumptensor and Ring. 
This implies that the retriever has a good chance to include the gold documentation within 4k context length; 
(2). LLMs have low attention in the middle of long contexts \cite{liu2023lost}.
With long contexts, LLMs may fail to identify the relevant content in the middle that can help solve the problem. 

We leave it to future research to design a more delicate retrieval system that can appropriately regulate the content utilized for LLM generation.

\section{Personally Identifying Infomation}
We collect data from the domain of code generation.
We authors carefully reviewed all the collected data, and confirm that the data that was collected/used does not contain any information that names or uniquely identifies individual people or offensive content.

\section{Intended use}
\ourmethod is an advanced pipeline for RACG, and it is expected to be applied in customized code generation.
\ourdata consists of four realistic benchmarks for RACG, and is expected to be used as an evaluation benchmark to evaluate RACG systems.

\section{License}
Our code and data will be released under Apache-2.0 license.

\section{Data Documentation}
\ourdata is collected from LeetCode and adapted from DS-1000~\cite{lai2023ds}, which was originally collected from StackOverflow.
The data in \ourdata is all in English.

\section{Machines}
We run all experiments on Nvidia A100 GPUs.
It takes around 1 hour to finish one dataset in \ourdata.
To complete all the experiments in the paper, it tasks around 24 hours.

\section{Packages}
For all the packages we use in the code, we employ the pip or conda implementation in the latest version.

\section{Instructions for Human Annotators}
\begin{itemize}
    \item Write the code in the corresponding programming languages to the problem
    \item Find the documentation of the syntax used in the code.
\end{itemize}

\section{Data Consent}
We adapt data from DS1000~\cite{lai2023ds}, which is released under CC-BY-SA-4.0 license.
Some of the data is collected from LeetCode, which is a public platform for practicing coding skills.
By the copyright law 107: Notwithstanding the provisions of sections 106 and 106A, the fair use of a copyrighted work, including such use by reproduction in copies or phonorecords or by any other means specified by that section, for purposes such as criticism, comment, news reporting, teaching (including multiple copies for classroom use), scholarship, or research, is not an infringement of copyright. In determining whether the use made of a work in any particular case is a fair use the factors to be considered shall include—
(1)the purpose and character of the use, including whether such use is of a commercial nature or is for nonprofit educational purposes;
(2)the nature of the copyrighted work;
(3)the amount and substantiality of the portion used in relation to the copyrighted work as a whole; and
(4)the effect of the use upon the potential market for or value of the copyrighted work.\footnote{https://copyright.gov/fair-use/}

We should be eligible to use the data.



%% file: tables/private_library.tex
\begin{table*}
\centering
\small
\begin{tabular}{lc cccc c cccc}
\toprule
 & & \multicolumn{4}{c}{\textbf{Model: ChatGPT}} & & 
\multicolumn{4}{c}{\textbf{Model: CodeLlama}}\\
\cline{1-1}
\cline{3-6}
\cline{8-11}
Method & & Monkey & BeatNum & TorchData & Avg. & & Monkey & BeatNum & TorchData & Avg. \\
\cline{1-1}
\cline{3-6}
\cline{8-11}
Vanilla & & 38.6 & 27.7 & 42.0 & 36.1 & & 80.2 & 70.3 & 54.0 &  68.2\\
\ourmethod & & \textbf{67.3} & \textbf{70.3} & \textbf{74.0} & \textbf{70.5} &  & \textbf{93.1} & \textbf{90.1} & \textbf{92.0} & \textbf{91.7} \\
\bottomrule
\end{tabular}

\caption{\label{table:private_lib} We evaluate the zero-shot ChatGPT and CodeLlama on three private libraries. 
Although we observe significant improvements of \ourmethod, the exceptionally high accuracy of CodeLlama Vanilla (zero-shot) performance suggests the risk of data leakage, making it less reliable to assess model generalization capabilities.
}
\end{table*}

%% file: tables/test_inputs.tex
\begin{table*}
\centering
\small
\begin{tabular}{cccc c cccc}
\toprule
\multicolumn{4}{c}{\textbf{Model: ChatGPT}} & & 
\multicolumn{4}{c}{\textbf{Model: CodeLlama}}\\
\cline{1-4}
\cline{6-9}
\newscience & Tensor-M & Ring & Pony & & \newscience & Tensor-M & Ring & Pony\\
\cline{1-4}
\cline{6-9}
89.2 & 93.3 & 100.0 & 100.0 & & 86.8 & 91.1 & 95.6 & 96.8\\
\bottomrule
\end{tabular}

\caption{\label{table:test_inputs}The accuracy of ChatGPT (left) and CodeLlama (right) in generating valid program inputs.
Although LLMs cannot guarantee to write accurate test cases, their performance in generating only program inputs is exceptionally high.
}
\end{table*}

%% file: tables/cost.tex
\begin{table*}[t]
\centering
\small
\begin{tabular}{cccccccccccccccc}
\hline
&  & \multicolumn{2}{c}{\newscience} & \multicolumn{2}{c}{\jumptensor}  & \multicolumn{2}{c}{Ring}  & \multicolumn{2}{c}{Pony}   & \multicolumn{2}{c}{Average}   \\
\cmidrule(lr){3-4} 
\cmidrule(lr){5-6}
\cmidrule(lr){7-8}
\cmidrule(lr){9-10}
\cmidrule(lr){11-12}
Model & Retrieval & Acc & Tokens & Acc & Tokens  & Acc & Tokens  & Acc & Tokens & Acc & Tokens   \\
\hline
\multirow{4}{*}{ChatGPT} &  Vanilla & 17.6 & 423 & 11.1 & 322 & 3.7 & 206  & 1.8 & 222 & 8.6 & 293\\
&  DocPrompting & 32.4 & 3687 & 33.3 & 3728  & 8.4 & 3826 & 2.7 & 3763 & 19.2 & 3751\\
&  RK & 32.6 & 4826 & 40.0 & 4924  & 11.7 & 4528  & 5.2 & 4584 & 22.4 & 4716\\
&  \ourmethod & 37.9 & 13631 & 53.3 & 12568  & 36.6 & 24987  & 13.5 & 13819 & 35.3 & 16251\\
\hline
\multirow{4}{*}{CodeLlama} &   Vanilla & 11.3 & 476 & 17.8 & 381  & 0.0 & 263  & 0.0 & 314 & 7.3 & 386 \\
&  DocPrompting & 16.9 & 3923 & 37.8 & 4012  & 4.7 & 4050  & 4.4 & 3978 & 16.0 & 3991\\
&   RK & 23.9 & 5124 & 42.2 & 5023  & 8.2 & 4987  & 7.3 & 4823 & 20.4 & 4989\\
&   \ourmethod & 31.2 & 14564 & 53.3 & 12323  & 26.7 & 29384  & 17.4 & 14592 & 32.2 & 17716\\
\hline
\end{tabular}
\caption{\label{tab:cost}The comparison of LLM performance and consumed tokens per example without retrieval (Vanilla), retrieval without evolution from only documentation (DocPrompting), retrieval without evolution from knowledge soup (RK) and \ourmethod retrieval.
By default, we use INSTRUCTOR-xl as the embedding model.
The results in the table demonstrate the association between superior results and the massively processed tokens, which implies the trade-off between the performance and the cost.
}
\end{table*}